\documentclass[runningheads]{llncs}
\usepackage{graphicx}
\usepackage{amsmath,amssymb} 
\usepackage{color}
\usepackage[width=122mm,left=12mm,paperwidth=146mm,height=193mm,top=12mm,paperheight=217mm]{geometry}

\usepackage{xcolor}
\usepackage{xspace}
\usepackage[pagebackref=true,breaklinks=true,letterpaper=true,colorlinks,bookmarks=false]{hyperref}
\usepackage{longtable}
\usepackage{subfigure}
\usepackage{dsfont}

\begin{document}
\pagestyle{headings}
\mainmatter
\title{Transferring Landmark Annotations for Cross-Dataset Face Alignment} 

\titlerunning{Transferring Landmark Annotations for Cross-Dataset Face Alignment}

\authorrunning{Zhu et al.}

\newcommand*\samethanks[1][\value{footnote}]{\footnotemark[#1]}
\author{Shizhan Zhu \inst{1}\thanks{denotes equal contributions.} \and Cheng Li \inst{2}\samethanks[1] \and Chen Change Loy \inst{1} \and Xiaoou Tang \inst{1}}
\institute{Department of Information Engineering, The Chinese University of Hong Kong \and %
           Department of Physics, Tsinghua University}

\newcommand*{\eg}{e.g.\@\xspace}
\newcommand*{\ie}{i.e.\@\xspace}
\newcommand*{\etc}{etc.\@\xspace}
\newcommand*{\etal}{et al.\@\xspace}

\maketitle

\begin{abstract}

Dataset bias is a well known problem in object recognition domain. This issue, nonetheless, is rarely explored in face alignment research.
In this study, we show that dataset plays an integral part of face alignment performance. Specifically, owing to face alignment dataset bias, training on one database and testing on another or unseen domain would lead to poor performance.
Creating an unbiased dataset through combining various existing databases, however, is non-trivial as one has to exhaustively re-label the landmarks for standardisation.
In this work, we propose a simple and yet effective method to bridge the disparate annotation spaces between databases, making datasets fusion possible.
We show extensive results on combining various popular databases (LFW, AFLW, LFPW, HELEN) for improved cross-dataset and unseen data alignment.

\keywords{Face alignment, dataset bias, transductive learning}
\end{abstract}

\section{Introduction} \label{sec_intro_}

Face alignment is a critical component of various face analyses, such as face verification~\cite{Berg_CVPR13,sun2013hybrid,sun2013deep}, face recognition~\cite{zhu2013deep}, age estimation~\cite{chen2013cumulative}, and expression classification~\cite{Wang_CVPR13}.
Various benchmark datasets~\cite{dataset_LFPW,dataset_LFW,dataset_AFLW,dataset_HELEN} have been released, each of which containing large quantities of labelled images.
Despite the databases were collected with the goal of being as rich and diverse as possible, inherent bias across datasets is unavoidable in practice~\cite{Torralba_CVPR11}.

\begin{figure}[h]
\centering
\includegraphics[width= 0.95 \linewidth]{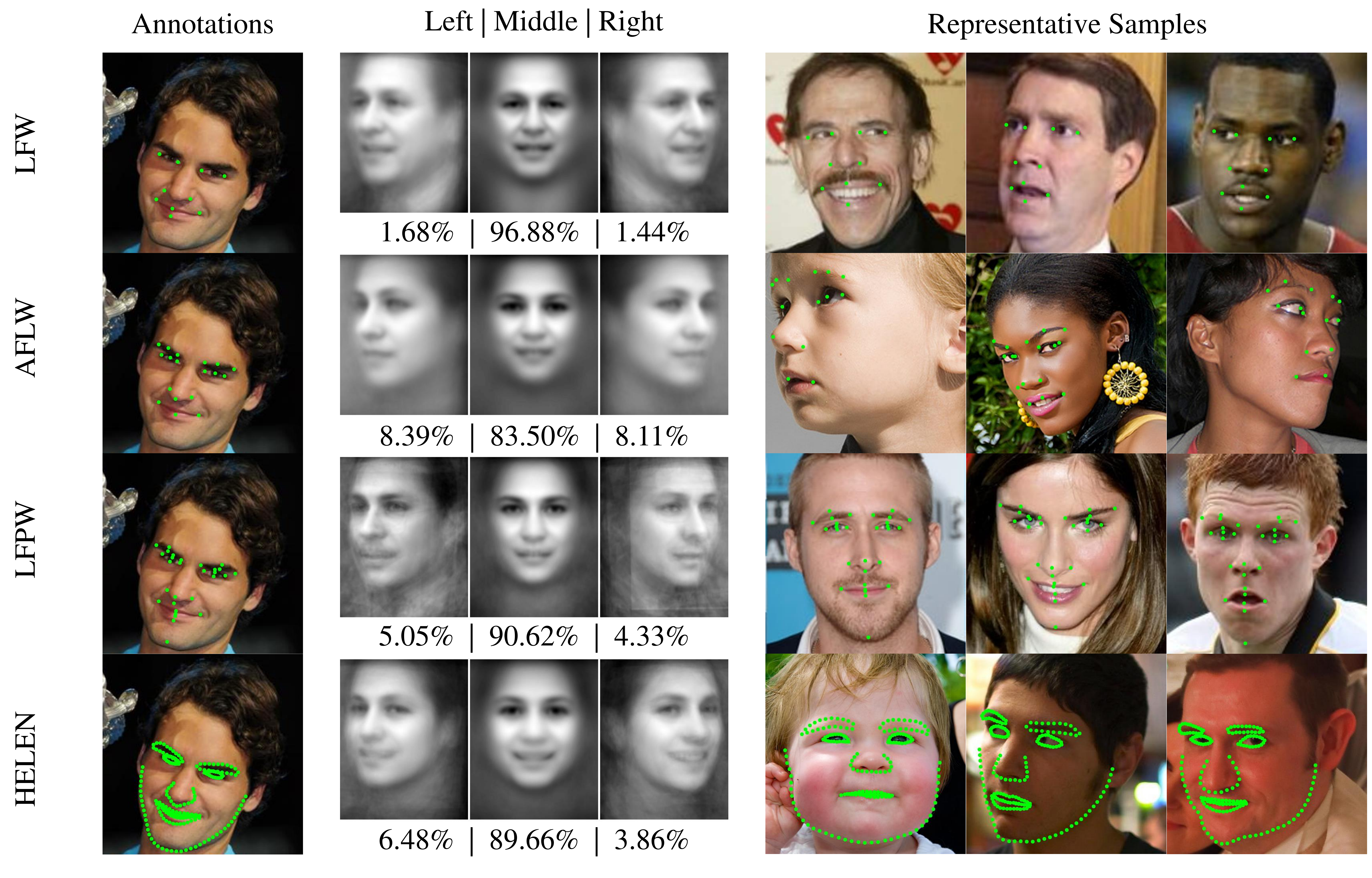}
\vskip -0.3cm
\caption{The differences between four popular face alignment datasets, \ie~LFW~\cite{dataset_LFW}, AFLW~\cite{dataset_AFLW}, LFPW~\cite{dataset_LFPW}, and HELEN~\cite{dataset_HELEN}. The left column shows the different annotation protocols, the middle column summarises the pose distributions, and the right column provides some representative examples of each dataset.}
\label{fig_intro_datasetsDifferences_}
\vspace{-0.45cm}
\end{figure}

The bias presents in the form of different characteristics and distributions across datasets, as depicted in Fig.~\ref{fig_intro_datasetsDifferences_}. For instance, one set mainly contains white Caucasian male with mostly frontal faces, while another set consists of challenging samples with various poses or severe occlusions.
In addition, the distribution difference between profile views can differ as much as over 10\% across datasets.
Clearly, training a model on one dataset would lead to over-fitting easily, and causing poor performance on unseen domain.
To improve generalisation, it is of practical interest to combine different databases so as to leverage the characteristics and distributions of multiple sources.
This thought, however, is hindered by the annotation gaps (see the first column of Fig~\ref{fig_intro_datasetsDifferences_}), which requires huge effort to standardize before databases fusion is possible.


The objective of this study is to formulate an approach that allows \textit{integration of different databases despite their different annotation protocols}.
At first glance, this seems unsolvable but we make this possible through exploiting \textit{common landmarks} across datasets.
Specifically, we observe that many landmarks are well labelled with decisive semantic definition across different datasets, \eg left and right eyes corners, mouth corners and pupil centers.
These common landmarks can usually be found on different datasets, although their numbers can be different. Often, there are 6 to 12 common landmarks annotated on a pair of datasets (the dark blue landmarks in Fig.~\ref{fig_method_tcr_pipeline_}).
Theses common landmarks provide us with an opportunity to transfer information from one dataset to the other.

To this end, we propose a simple yet effective approach to exploit common landmarks as guidance, and transfer labelled landmarks from a given source data and fitting them to the images in an arbitrary target set.
%
%
%
%
%
%
%
Performing datasets fusion with the proposed approach offers us with enormous advantages:
(i) \textit{sample diversification} -- our method allows standardisation of disparate annotation spaces, thus allowing fusion of datasets. The combined dataset captures the diverse characteristics of samples from multiple sources. A model trained on this dataset is expected to have better generalisation on unseen samples.
(ii) \textit{annotation enrichment} -- we are no longer limited to models only capable of particular landmark configuration if it is trained on a specific dataset. With the proposed approach, one can transfer densely labelled annotations from a source to sparsely labelled target for high-quality dense annotation in the target domain.



\vspace{0.1cm}
\noindent \textbf{Contribution:}
We show for the first time how annotation spaces of different face alignment datasets can be standardised automatically. This allows us to combine and exploit diverse datasets for training a single model. Extensive experiments show that the resulting model achieves state-of-the-art face alignment results in cross datasets and unseen domain data evaluations.
In particular, we achieve \textbf{16.6}\% improvement on average against the method with `closed-world' assumption when performing cross-datasets evaluations, and \textbf{11.4}\% improvement on average compared to na\"{i}ve training sets fusion.
Based on the proposed annotation transferring approach, we obtain and release dense annotations (68 and 194 points) on the popular face verification dataset LFW \cite{dataset_LFW} via the link \url{http://mmlab.ie.cuhk.edu.hk/projects/landmarksTransferring.html}.


\section{Related Work} \label{sec_related_}

\textbf{Face Alignment} Approaches for face alignment can be broadly divided into three categories: (i) active appearance model based method, (ii) cascaded regression method, and (iii) detection based method.
As the most classic method, the original \textsl{active appearance model} (AAM)~\cite{Cootes_ECCV98} tries to search for shape parameters through minimising the residual between the face appearance and a face template. The method suffers from poor generalisation and sensitivity to initialisation.

\textsl{Cascaded regression} treats shape estimation as a regression problem. It starts from raw estimates of landmark positions, and learns regressors that map shape dependent features into pose increments iteratively.
Examples of cascaded regression method include the approach by Cao~\etal~\cite{Cao_CVPR12}, which employs boosted nonlinear regression with shape dependent pixel difference features.
Burgos-Artizzu~\etal~\cite{Burgos-Artizzu_ICCV13} builds a cascaded regression model with an occlusion detection and voting strategy to cope with severe occlusion.
Xiong and De la Torre~\cite{Xiong_CVPR13} address regression by learning generic descent directions, and perform linear mapping on non-linear SIFT features, which achieves state-of-the-art results.
%

\textsl{Detection based approaches} detects object parts independently and then estimates pose and/or shape directly from the detections~\cite{Cevikalp_FG13,Cootes_ECCV12} or through flexible part models~\cite{Burl_ECCV98,Bourdev_ICCV09,Sande_PAMI10}.
These methods are effective at detecting and localising articulated objects from multiple views in challenging scenarios.
Sun~\etal~\cite{Sun_CVPR13} propose a cascaded deep convolutional network for five-points face alignment.
The network detects approximate locations of the landmarks in the lower cascade and refine the estimations in higher cascade.
State-of-the-art result is recently achieved by Zhang~\etal~\cite{zhang2014facial,zhang2014learning} using a deep convolutional network trained for facial landmark detection together with heterogeneous but subtly correlated tasks, like head pose estimation and facial attribute inference. The model achieves the state-of-the-art result on the 300-W benchmark dataset (mean error of 9.15\% on the challenging IBUG subset).


\vspace{0.1cm}
\noindent \textbf{Dataset Bias} Torralba and Efros~\cite{Torralba_CVPR11} raise an important question: are the datasets deployed for computer vision studies unbiased representations of the visual world?
They showed that even large number of training images are employed, an image classification model can still over-fit if it is trained on a single dataset with bias. The over-fitting would severely hamper cross-dataset generalisation.
A number of studies focus on undoing this bias by transfer learning~\cite{Saenko_ECCV10,Kulis_CVPR11,Tommasi_ACCV12} or through other means like max-margin based learning framework or subspace alignment method~ \cite{Khosla_ECCV12,Fernando_ICCV13}.
To our knowledge, our work is among the first studies that investigates the problem of dataset bias in face alignment domain.
We wish to show that using existing databases independently for training/test would risk a `closed-world' evaluation environment. To allow improved cross-dataset generalisation, we devise a novel transductive alignment method to bridge the annotation gap between diverse datasets, which in turn facilitates seamless databases fusion for domain adaptive face alignment.

It is worth noting that in the recent work by Smith and Zhang~\cite{Smith_ECCV14}, they have independently presented an alternative way to combine multiple face landmark datasets with different landmark definitions into a super dataset.

\section{Methodology} \label{sec_method_}
\vspace{-0.15cm}
\subsection{Problem and Notations} \label{sec_method_notations_}
In a typical face alignment pipeline, one assumes the training set $\mathcal{D}_{train} = $ $\{( \mathbf{x}^\ast , I)\}$ consists of both images $I$ and the corresponding ground truth coordinates $\mathbf{x}^\ast$, each image $I_i$ contains a cropped face and each ground truth pose $\mathbf{x}^\ast_i \in \mathds{R}^{2 \times n}$, in which $n$ is the number of landmarks on each face. We use asterisk to denote ground-truth coordinates.

The goal is to learn a model $\Theta_\mathcal{D}$, to estimate the location of landmarks in the test set $\mathcal{D}_{test}$.
%
%
For a cascaded regression method, the estimate of the current landmarks' coordinates for iteration $k$ is denoted by $\mathbf{x}_k$. For clarity, we use $\mathbf{x}$ for abbreviation in the following discussion.
We use $\phi(\mathbf{x})$ to represent the shape-indexed features extracted according to the specific pose parameterised by $\mathbf{x}$. 
%
%
For a SIFT-based shape dependent features~\cite{Xiong_CVPR13}, the dimension of the features $\phi$ is $n \times 128$.

\begin{figure}
\centering
\includegraphics[width=0.9\linewidth]{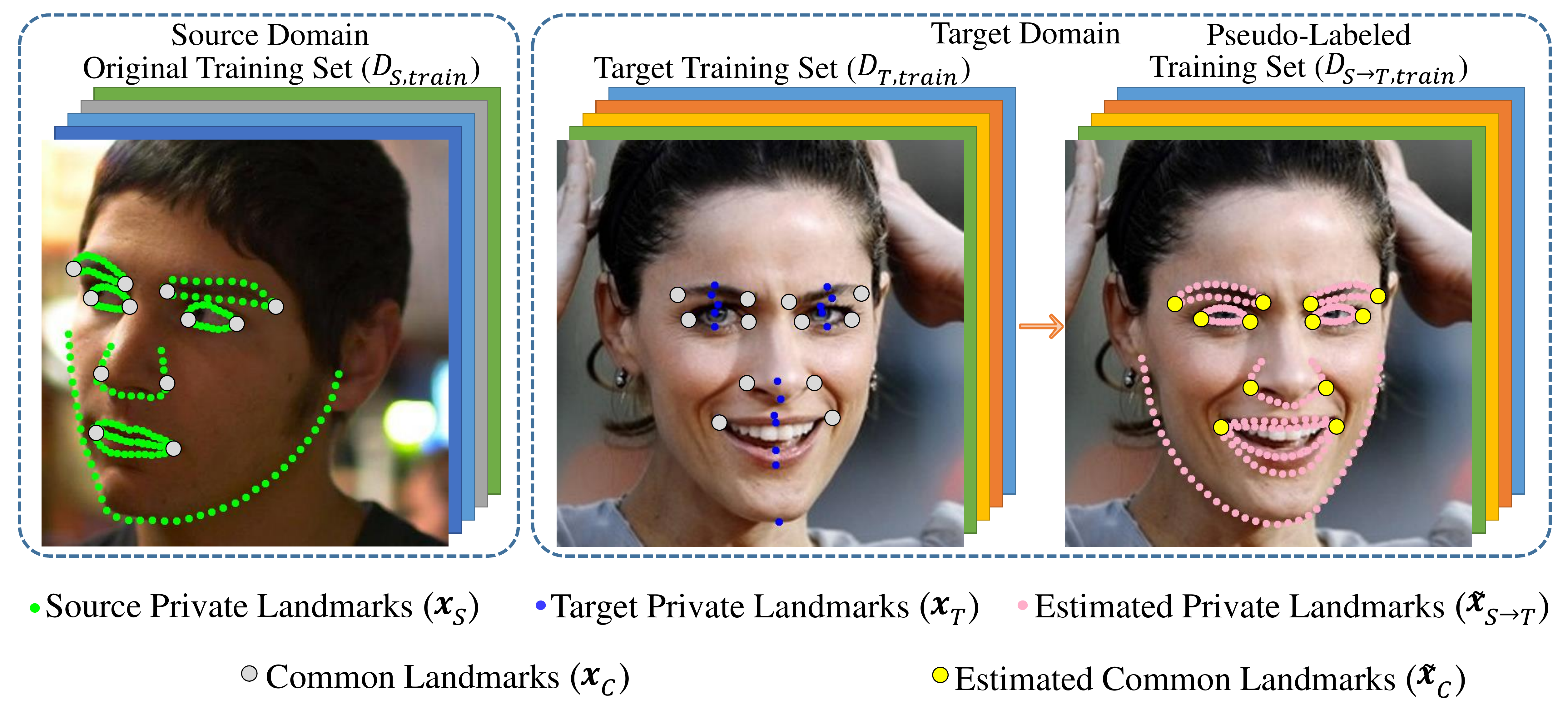}
\vskip -0.25cm
\caption{Relations between notations. Best viewed in colour. Note that the existence of notation $\ast$ indicates ground truth. If the landmarks on the faces are human annotations, these notations all need to add $\ast$.}
\label{fig_method_notations_notations_}
\end{figure}

In this study, we assume there exists a source dataset, represented as $\mathcal{D}_{S,train} = \{((\mathbf{x}^\ast_C,\mathbf{x}^\ast_S)^\top,I_S)\}$ for training and $\mathcal{D}_{S,test}$ for testing. On the other hand, we have a target set $\mathcal{D}_{T,train} =  \{((\mathbf{x}^\ast_C,\mathbf{x}^\ast_T)^\top,I_T)\}$ and $\mathcal{D}_{T,test}$.
More precisely, as shown in Fig.~\ref{fig_method_notations_notations_}, the source and target training sets share some \textit{common landmarks} $\mathbf{x}^\ast_C$, which co-exist between source and target training sets.
On the other hand, there exist \textit{private landmarks}, which refer to those that can only be found on either source or target training sets, but not both. They are represented as $\mathbf{x}^\ast_S$ and $\mathbf{x}^\ast_T$, respectively.

Note that despite the source and target training sets share some common landmarks, their private landmarks are different, thus the total number of landmark annotations, $n_S$, and $n_T$, are also different. 
Our task is to bridge such annotation gap. 
As discussed in Section~\ref{sec_intro_}, performing such an annotation transfer operation is challenging, in that the annotation protocols of source and target sets could differ significantly.
We address this problem through exploiting the common landmarks co-exists between the source and target sets. The details are presented in Section~\ref{sec_method_ta_}.

%
%


Transferring the annotations from source to target will provide the target set with \textit{source-type landmarks}, as shown in the right-most subfigure in Fig.~\ref{fig_method_notations_notations_}.
The transferred private landmarks from the source to the target set are denoted as $\tilde{\mathbf{x}}_{S\rightarrow T}$, whilst the transferred common landmarks are given as $\tilde{\mathbf{x}}_{C}$.
The target set with this new set of annotations is known as pseudo-labeled target training set, and it is represented as $\mathcal{D}_{S\rightarrow T,train} = \{( (\tilde{\mathbf{x}}_{C}, \tilde{\mathbf{x}}_{S\rightarrow T})^\top ,I_T)\}$.
%
%
We show in Section~\ref{sec_exp_ta_} that the transferred annotations/landmarks are close to human annotating accuracy.
We can readily combine the synthesised target training set with the source training set, since they now have an identical set of annotations. We show this possibility in Section~\ref{sec_method_tcr_}.




\subsection{Brief Overview of Supervised Descent Method} \label{sec_method_overview_}

Before we detail how annotation transfer is performed, we first provide a brief review on the Supervised Descent Method (SDM)~\cite{Xiong_CVPR13} as it forms the basis of the proposed approach. We note that our concept of transferring annotations is not limited to the SDM method but can be adapted to other existing cascaded regression-based approaches~\cite{Cao_CVPR12,Burgos-Artizzu_ICCV13}.


In SDM, faces are centered to a mean shape and initial poses for all samples are initialised as the mean pose. Features are then extracted from the initial landmarks.
%
%
The goal of SDM is to refine the current landmarks' locations iteratively following a movement target.
%
%
This is achieved by defining a loss function
\begin{equation}
\label{eqn:orig_loss}
L(\mathbf{x} + \Delta \mathbf{x}) = \| \phi(\mathbf{x}+\Delta \mathbf{x}) - \phi(\mathbf{x}^*) \|_2^2.
\end{equation}
The movement target $\Delta \mathbf{x}$ can be obtained through optimisation method by linear regression.
Here the loss function can be approximated by its second order Taylor expansion $L(\mathbf{x} + \Delta \mathbf{x}) \approx L(\mathbf{x}) + J_L(\mathbf{x})^\top\Delta \mathbf{x} + \frac{1}{2}\Delta \mathbf{x}^\top H_L(\mathbf{x}) \Delta \mathbf{x}$, where $J_L$ and $H_L$ are Jacobian and Hessian matrix of the loss function $L$, respectively. The approximate solution for $\Delta \mathbf{x}$ can be obtained by $\Delta \mathbf{x} = -2H_L^{-1}J_\phi(\phi(\mathbf{x}) - \phi(\mathbf{x}^*))$.
%

Since features are not always differentiable and performing numerical differential is computationally expensive, the projection from derivatives of features to $\Delta \mathbf{x}$ is estimated through the following regression problem:

\begin{equation} \label{eqn1}
\Delta \mathbf{x} = \mathbf{A}(\phi(\mathbf{x}) - \phi(\mathbf{x}^*)),
\end{equation}
where $\mathbf{A}$ is the coefficient matrix.
However, in the testing stage, the specific locations of ground truth $\mathbf{\mathbf{x}}^\ast$ is not available, thus it is impossible to extract features $\phi(\mathbf{x}^*)$.
The SDM resorts to approximation by using $\phi(\mathbf{x})$ to replace the factor $(\phi(\mathbf{x}) - \phi(\mathbf{x}^*))$ so that the regression can be learned as follows:

\begin{equation} \label{eqn_approximation}
\Delta \mathbf{x} = \mathbf{A} \phi(\mathbf{x}) + \mathbf{b}
\end{equation}

This formulation inspires the proposed transductive alignment method, which will be presented next. In particular, in cross-dataset annotation transfer, we could actually exploit the set of common landmarks to estimate $\phi(\mathbf{x}^*)$.
Given the estimated $\phi(\mathbf{x}^*)$, we could achieve better regression performance than approximation as in Eq.~\ref{eqn_approximation}.
%


\subsection{Transductive Alignment} \label{sec_method_ta_}

The core step of our approach is transductive alignment.
The goal of transductive alignment is to obtain the synthesized target training set $\mathcal{D}_{S\rightarrow T,train} = \{( (\tilde{\mathbf{x}}_{S\rightarrow T}, \tilde{\mathbf{x}}_{C})^\top ,I_T)\}$, as shown in Fig.~\ref{fig_method_notations_notations_}.
We will obtain the transferred annotations $(\tilde{\mathbf{x}}_{S\rightarrow T}, \tilde{\mathbf{x}}_{C})^\top$ with the guidance of common landmarks $\mathbf{x}_C^*$. The details are given as follows.



%


To extend the original SDM for transductive alignment, we need to estimate those unknown features $\phi(\mathbf{x}^*)$. In SDM, all information from $\phi(\mathbf{x}^*)$ is counted implicitly into the bias term, causing a loss in information.
Considering our task of forming synthesized training dataset $\mathcal{D}_{S\rightarrow T,train} = \{( (\tilde{\mathbf{x}}_{S\rightarrow T}, \tilde{\mathbf{x}}_{C})^\top ,I_T)\}$, we actually have extra information from common landmarks $\mathbf{x}_{C}^*$. Here we attempt to partially recover the missing term $\phi(\mathbf{x}^*)$ in Equation~\ref{eqn_approximation} by $\phi(\mathbf{x}_{C}^*)$.
Since the features extracted around the landmarks always share a considerable extent of overlapping area, especially in densely annotated region, features $\phi(\mathbf{x}_{C}^*)$ and $\phi(\mathbf{x}_{S}^*)$\footnote{We denotes $\mathbf{x}_{S}^*$ as $\mathbf{x}^* \setminus \mathbf{x}_{C}^*$, thus $\phi(\mathbf{x}_{S}^*) = \phi(\mathbf{x}^*) \setminus \phi(\mathbf{x}_{C}^*)$.} thus have high correlation.

More precisely, we assume we can estimate $\phi(\mathbf{x}^*_S)$ by a linear projection from $\phi(\mathbf{x}^*_{C})$, as
\begin{equation}
\tilde{\phi}(\mathbf{x}_S^*) = f(\phi(\mathbf{x}_{C}^*)) = \mathbf{W} \phi(\mathbf{x}_{C}^*) + \mathbf{t}
\end{equation}
Substituting this back to Equation~\ref{eqn1}, we found that $\Delta x$ also has linear relation with $(\phi(\mathbf{x}_S^*),\phi(\mathbf{x}_{C}^*))^\top$.
Since we add relatively accurate $\tilde{\phi}(\mathbf{x}^*_S)$ estimations for the regression, it would be more suitable if we apply the following regression strategy:
\begin{equation} \label{eqn_semi_reg}
\begin{aligned}
\left[
  \begin{array}{c}
    \Delta \mathbf{x}_C\\
    \Delta \mathbf{x}_S\\
  \end{array}
\right]
= & \mathbf{A}
\left[
  \begin{array}{c}
    \phi(\mathbf{x}_C)\\
    \phi(\mathbf{x}_S)\\
    \phi(\mathbf{x}_C^*)\\
    \phi(\mathbf{x}_S^*)\\
  \end{array}
\right]
+\mathbf{b} 
= [\mathbf{A}, \mathbf{b}]
\left[
   \begin{array}{c}
    \phi(\mathbf{x}_C)\\
    \phi(\mathbf{x}_S)\\
    \phi(\mathbf{x}_C^*)\\
    \phi(\mathbf{x}_S^*)\\
    1
  \end{array}
\right] \\
= & [\mathbf{A}, \mathbf{b}]
\begin{bmatrix}
\mathbf{I} & & &\\
& \mathbf{I} & & \\
& & \mathbf{I} & \\
& & \mathbf{W} & \mathbf{t} \\
& & & 1
\end{bmatrix}
\left[
   \begin{array}{c}
    \phi(\mathbf{x}_C)\\
    \phi(\mathbf{x}_S)\\
    \phi(\mathbf{x}_C^*)\\
    1
  \end{array}
\right] \\
= & [\mathbf{A}', \mathbf{b}']
\left[
   \begin{array}{c}
    \phi(\mathbf{x}_C)\\
    \phi(\mathbf{x}_S)\\
    \phi(\mathbf{x}_C^*)\\
    1
  \end{array}
\right] 
= \mathbf{A}'
\left[
  \begin{array}{c}
    \phi(\mathbf{x}_C)\\
    \phi(\mathbf{x}_S)\\
    \phi(\mathbf{x}_C^*)\\
  \end{array}
\right]
+\mathbf{b}' \\
\end{aligned}
\end{equation}

%
where $\mathbf{x}_C,\mathbf{x}_S$ are the estimated source-type common and private landmarks, whilst $\mathbf{x}_C^\ast$ is the ground-truth common landmarks. $\mathbf{A}'$ and $\mathbf{b}'$ denote the regression coefficient matrix and bias learned using the source dataset.

\begin{figure}%
\centering
\includegraphics[width=0.8\linewidth]{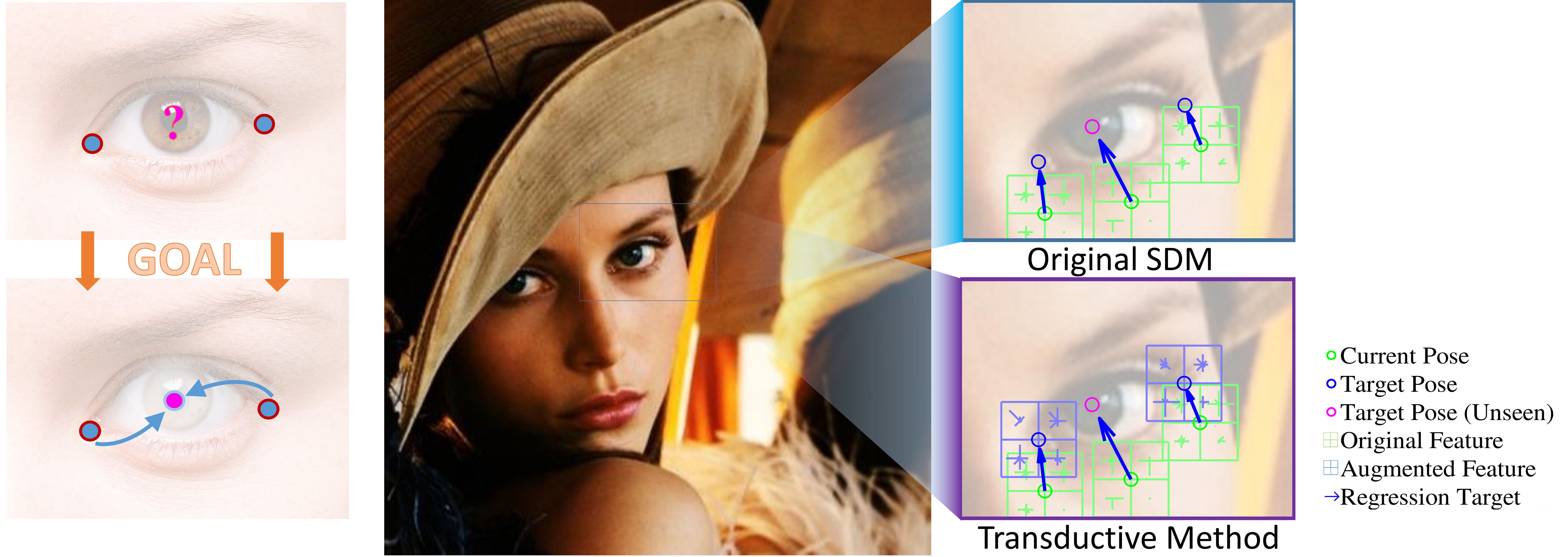}
\caption{Basic idea of the proposed transductive alignment. The goal is to automatically label source-type landmarks (\textcolor{magenta}{magenta}) under the guidance of common landmarks (\textcolor{blue}{blue}).}
\label{fig_method_ta_semi_}
\end{figure}

Figure~\ref{fig_method_ta_semi_} summarizes our transductive alignment step in an intuitive schematic diagram.
We obtain substantial improvement using the reference information from features $\phi(\mathbf{x}^\ast_C)$ extracted from common landmarks. Note that we do not directly use the specific location information from $\mathbf{x}^\ast_C$ in our estimation mainly because we need to prevent improvement bias, in which estimate on $\mathbf{x}^*_C$ improves a great deal, while $\mathbf{x}^*_S$ is still in poor modification, since $\phi(\mathbf{x}^\ast_C)$ provides global reference information beneficial to all the landmarks but $\mathbf{x}^*_C$ only contributes to itself. Experiments in Section \ref{sec_exp_ta_} demonstrate that the proposed transductive alignment method produces accurate source-type annotations on target domain.

\subsection{Augmenting Source and Target Training Sets} \label{sec_method_tcr_}

\begin{figure}
\centering
\includegraphics[width= 0.85 \linewidth]{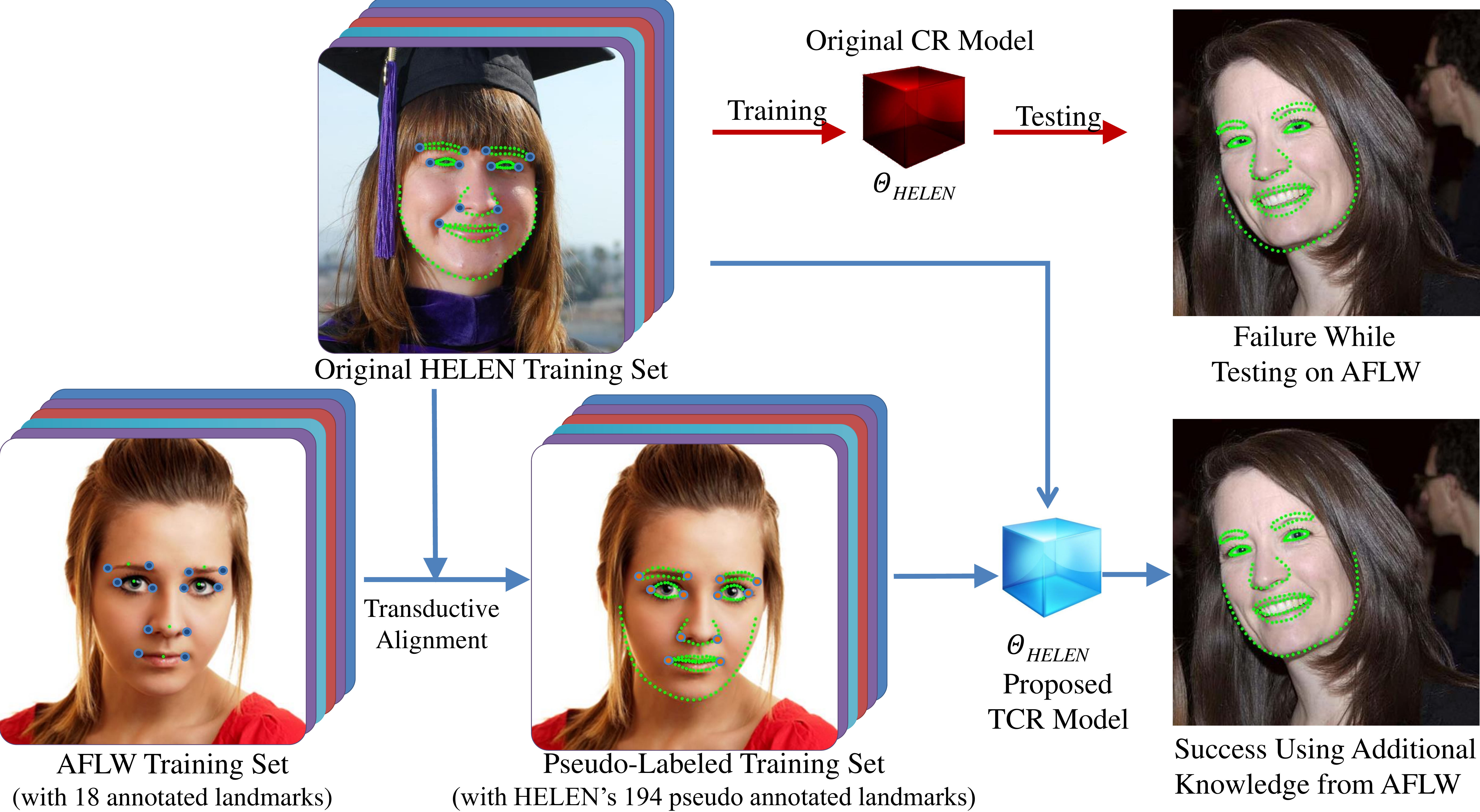}
\caption{The pipeline of the proposed Transductive Cascaded Regression (TCR). We show example of transferring annotations from the source HELEN~\cite{dataset_HELEN} into the target AFLW~\cite{dataset_AFLW}. Note that the annotations (green points) on the target training set is automatically labelled by our transductive alignment algorithm. Points in large size are common landmarks, labelled in both the source and target training datasets. The typical `closed-world' cascaded regression (CR) method is also drawn for comparison (depicted with red arrows).}
\label{fig_method_tcr_pipeline_}
\end{figure}

%
%
Figure~\ref{fig_method_tcr_pipeline_} shows the full pipeline of our proposed algorithm, including the step of augmenting the source and pseudo-labeled target training sets. We call the full pipeline as Transductive Cascaded Regression (TCR).
\begin{itemize}
\item \textbf{Step 1} -- Unlike the conventional cascaded model learning process (depicted with red arrows), we first obtain the pseudo-labeled target training set $\mathcal{D}_{S\rightarrow T,train}$ by transductive alignment described in Section \ref{sec_method_ta_}.
\item \textbf{Step 2} -- We then filter erroneous transferred annotations in the pseudo-labeled target training set. This is done through comparing the estimated $\tilde{\mathbf{x}}_{C}$ and ground truth $\mathbf{x}_C^\ast$ common landmarks. In particular, we remove target training samples with  error larger than $\epsilon$ in their estimated common landmarks. Only those samples with valid transferred annotations remain in the pseudo-labeled target training set.
The filtered transferred annotations are clean and close to human annotation, as we will show in Section~\ref{sec_exp_ta_}.
\item \textbf{Step 3} --  We combine the cleaned pseudo-labeled target training set, $\mathcal{D}_{S\rightarrow T,train}$, with the source training set $\mathcal{D}_{S, train}$.
\item \textbf{Step 4} --  A model is learned using the combined training set.
\end{itemize}

Note that Step 3 is possible thanks to the transductive alignment step, which bridge the annotation gap between the source and target training sets.
Next, we demonstrate the effectives of the proposed approach in cross-dataset and unseen data evaluation, and its robustness in handling challenging settings, such as large pose variations and severe occlusions.

\section{Experimental Settings} \label{sec_settings_}

\noindent \textbf{Datasets}: We selected a number of popular face alignment datasets for evaluation. These datasets are different in terms of their distributions in pose variations, and the degrees of illuminance and occlusion. Table~\ref{tab_settings_dataset_} summarizes the datasets, with sample images provided in Fig.~\ref{fig_intro_datasetsDifferences_}.

\begin{table}[htb]
	\centering
	\caption{ Summary of datasets, \textbf{\#images}: number of images in the original dataset. \textbf{\#train/test} : number of training/test images we use. \textbf{pts}: number of labelled landmarks. \textbf{\#miss-detect}: number of images missed by the detector~\cite{multi-view_detector} we employ. For LFPW, some images are not available due to obsolete URL. For AFLW, we deleted some samples unsuited for SDM (\eg~with one eye totally unseen). For LFW and AFLW, we randomly selected 20\% samples of the sets for testing. }
	\vspace{-0.2cm}
	\label{tab_settings_dataset_}
	\begin{tabular}{ c | c c c c | l }
	\hline
	Name & \#images & \#train/test  & pts & \#miss-detect & Characteristics \\ \hline
    LFW\cite{dataset_LFW} & 13233 &  10389/2595	 & 10  & 249 & Mainly male with frontal faces \\ \hline
    AFLW\cite{dataset_AFLW} & 21123 &  9565/2396	& 18 & 178 & Challenging in pose variation  \\ \hline
    LFPW\cite{dataset_LFPW} & 1432 &  782/188	& 29 & 0 & Faces are mostly frontal \\ \hline
    HELEN\cite{dataset_HELEN} & 2330 &  2000/330	& 194 & 0 & Extreme closeup \& dense labels \\ \hline
    \end{tabular}
    \vspace{-0.25cm}
\end{table}

These four datasets can be combined differently to form a source and target pairs, resulting into  12 possible combinations. Training on these combinations gives us 12 models for cross-datasets evaluation.
Recall that we aim to predict source-type landmarks on target testing set. We therefore require extra labelling to generate ground truth for evaluation.
We collected all four types of annotations (LFW 10 pts, AFLW 18 pts, LFPW 29 pts, HELEN 194 pts) on HELEN and LFPW.
In addition, we also selected 40 samples randomly from the testing sets of LFW and AFLW with challenging pose variations, and labelled them manually with two other types of annotations\footnote{We do not label LFW and AFLW with 194 landmarks since we do not have the special annotation tool~\cite{dataset_HELEN}.} to form testing sets LFW-C and AFLW-C.
Note that all the additional labelled landmarks are only used for evaluation purpose.

\vspace{0.1cm}
\noindent \textbf{Performance Evaluation}: Similar to previous studies~\cite{Cao_CVPR12,Xiong_CVPR13,Burgos-Artizzu_ICCV13}, we measured error as the Root Mean Square Error (RMSE) percentage of the interocular distance. Estimations with error larger than 10\% are reported as failure cases~\cite{Burgos-Artizzu_ICCV13}.

\noindent \textbf{Comparison}: To our knowledge, no method exists for transferring annotations to perform cross-dataset face alignment. Thus no suitable baselines can be found.
%
%
We choose SDM \cite{Xiong_CVPR13} as our baseline method, because it's the most closest to our approach which can be compared under same initialization and feature settings.
It achieves state-of-the-art performances on most of the popular benchmark datasets. However, it does not have the capability to exploit additional dataset due to annotation discrepancy.
Since the training codes for~\cite{Xiong_CVPR13} is not publicly available, and most of the employed databases are not shared, we re-implemented the method, and verified the correctness of our implementation on LFW and HELEN.
%


\vspace{0.1cm}
\noindent \textbf{Implementation Details}: In our framework, faces were detected using a multiview Viola-Jone detector~\cite{multi-view_detector}, which returned not only a bounding-box for each face, but also a rough pose category label. Number of miss-detected images in each dataset is reported in Table~\ref{tab_settings_dataset_}. We initialized the face by aligning faces to a mean pose in a 250 $\times$ 250 normalised square. We set initialized landmarks for Step 1 as mean locations of all samples in training set.
We tuned parameters following the same settings as in~\cite{Xiong_CVPR13}: training samples were perturbed 10 times by a random rigid transform, and we reduced the dimensionality of the regression data by performing PCA preserving 98\% of the energy of the extracted features. We used SIFT features with fixed direction and each descriptor covers 20 $\times$ 20 pixels.
%
In the 12 pairs of cross-dataset evaluations, the number of common landmarks ranges from 6 to 12.
The threshold error $\epsilon$ for selecting valid pseudo annotation is fixed at 7.5 throughout the experiments.

\section{Results} \label{sec_exp_}
\vspace{-0.1cm}
\subsection{Evaluating the Effectiveness of Transductive Alignment} \label{sec_exp_ta_}


In this experiment, we wish to verify the effectiveness of transductive alignment method (Step-1 in Sec.~\ref{sec_method_tcr_}) by evaluating the accuracy of transferred annotations on the target training set.
Specifically, the evaluations were conducted to measure the mean error (i) the transferred common landmarks, and (ii) the transferred private landmarks, onto the target training set.
%
%
Since only the LFPW and HELEN datasets have all four types of ground truth annotations (see Sec.~\ref{sec_settings_}), our evaluation was limited to 6 source-target pairs.
The average errors of transferred common landmarks and transferred private landmarks are \textbf{3.35} and \textbf{3.87}, respectively.
Both errors are very close to human annotation performance, which is commonly within the range between 3.0 and 4.5 \cite{Burgos-Artizzu_ICCV13}.

\begin{figure}%
\centering
\includegraphics[width=0.85\linewidth]{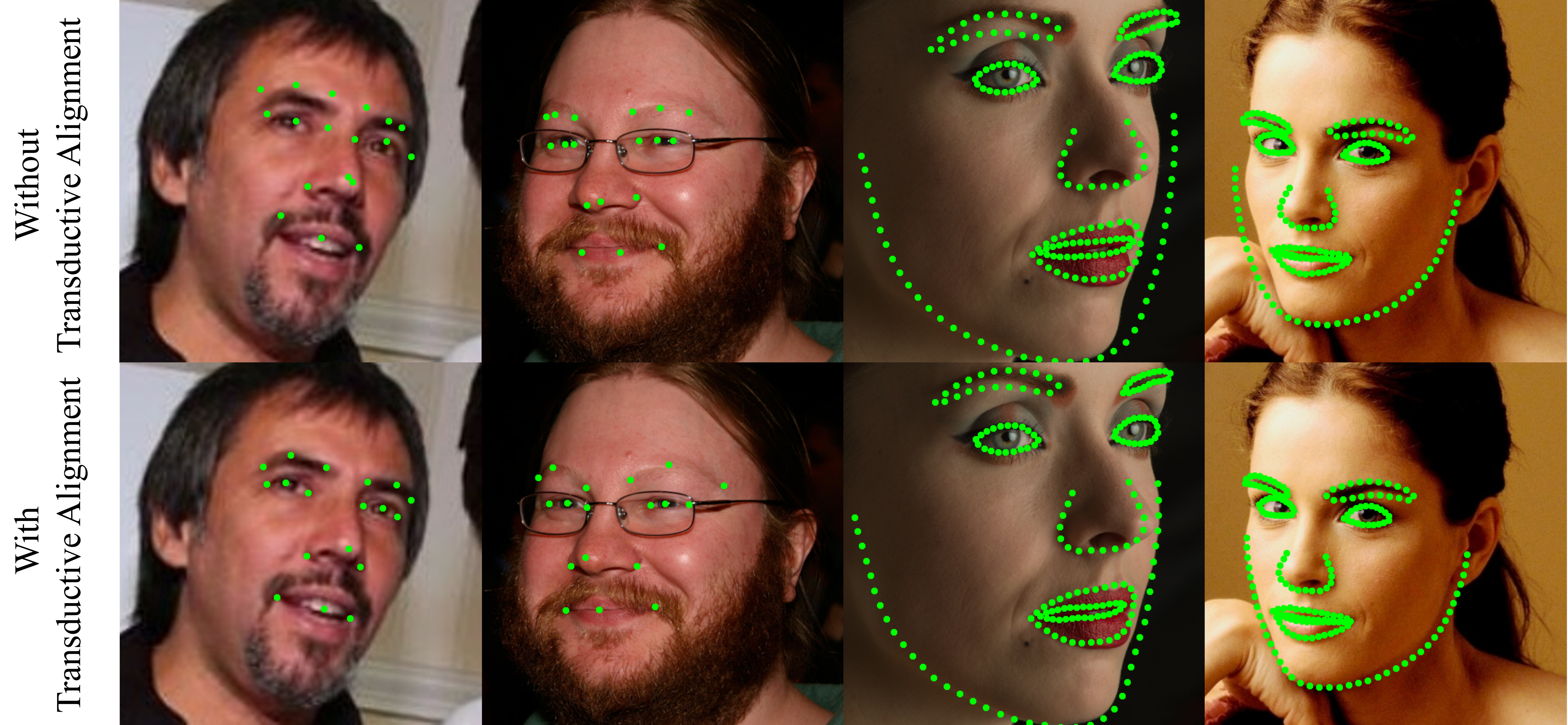}
\caption{Comparing the target training set annotations, generated by (top row) a naive cross dataset labelling strategy, and (bottom row) the proposed transductive alignment approach. Our methods outperforms due to the extra guidance of common landmarks co-exist between the source and target training sets.}
\label{fig_exp_ta_semiResult_}
\vspace{-0.25cm}
\end{figure}

Note that a straightforward way to label the target training set is to train a model on source training set and apply the model to infer landmarks on the target.
Figure~\ref{fig_exp_ta_semiResult_} shows the results obtained with such a na\"{i}ve cross-dataset alignment, and compares with that yielded by the proposed transductive alignment method.
Clearly, our approach outperforms, thanks to the guidance offered by the common landmarks.


\subsection{Evaluating TCR on Common Landmarks} \label{sec_exp_common_}

In this experiment, we evaluate the performance of our full Transductive Cascaded Regression model (the training steps described in Sec.~\ref{sec_method_tcr_}).
We compared the proposed model against the state-of-the-art method SDM~\cite{Xiong_CVPR13}. We used it to represent a `closed-world' method without annotation transfer and source-target sets augmentation.
In addition, we also compared against a na\"{i}ve fusion method, which simply combines the source and target sets without transductive alignment. In particular, this na\"{i}ve method was only trained with common landmarks from the source and target training sets. It therefore can predict only common landmark locations during testing.
Note that although both our method and SDM are capable of predicting full source-type landmarks, we only used them for predicting common landmarks.



Table~\ref{tab_exp_common_results_} summarizes the results.
In general the proposed TCR outperforms the two baselines.
Several observations are outlined below:
%
%
\begin{enumerate}
  \item From the diagonal values under the SDM column, we can observe that the alignment result is reaching its best when the training and testing are conducted on the same dataset. Nevertheless, the performance deteriorates in cross-dataset evaluations. These results strongly suggests the existence of dataset bias~\cite{Torralba_CVPR11}.
  \item The proposed TCR method obtains superior performance over the SDM, which assumes a `closed world' training/test environment. 
  \item The proposed model also outperforms the na\"{i}ve fusion method in most of the source-target pairs. The reason is that our model learns from a richer set of landmarks transferred from the source domain. In comparison, the na\"{i}ve fusion method only learns from common landmarks alone. The richer set of landmarks offers more constrains to the face shape, which in turn leads to more accurate estimation~\cite{Cootes_ECCV98}. This highlights the values of performing transductive alignment for annotation enrichment.
\end{enumerate}



\vspace{-0.5cm}
\begin{table}[htb]
  \centering
  \scriptsize
  \caption{Average error (\% interocular distance) for common landmarks. The best performance of each pair is shown in bold font. $S=$ source, $T=$ target.}
    \begin{tabular}{ l |c c c c || c c c c || c c c c }
    \hline
    \multicolumn{1}{r|}{$S$} & \multicolumn{4}{c||}{Closed-World (SDM \cite{Xiong_CVPR13})} & \multicolumn{4}{c||}{Na\"{i}ve Training Set Fusion} & \multicolumn{4}{c}{The Proposed TCR}\\
     \multicolumn{1}{l|}{$~T$} & LFW  & AFLW & LFPW & HELEN & LFW  & AFLW & LFPW & HELEN  & LFW  & AFLW & LFPW & HELEN \\\hline
    LFW & \textcolor{lightgray}{4.02} & 6.99 & 6.07 & 6.89 & \textcolor{lightgray}{-} & 5.33 & 5.10 & \textbf{5.40} & \textcolor{lightgray}{-} & \textbf{4.47} & \textbf{4.67} & 5.48 \\\hline
    AFLW & 8.88 & \textcolor{lightgray}{7.45} & 9.16 & 10.35 & 8.01 & \textcolor{lightgray}{-} & 8.56 & \textbf{9.15} & \textbf{7.31} & \textcolor{lightgray}{-} & \textbf{8.30} & 9.23 \\\hline
    LFPW & 5.52 & 6.43 & \textcolor{lightgray}{3.87} & 6.91 & 5.03 & 6.13 & \textcolor{lightgray}{-} & 6.09 & \textbf{3.69} & \textbf{4.97} & \textcolor{lightgray}{-} & \textbf{5.48} \\\hline
    HELEN & 8.07 & 8.94 & 6.93 & \textcolor{lightgray}{5.44} & 5.93 & 7.23 & 6.89 & \textcolor{lightgray}{-}  & \textbf{4.83} & \textbf{5.93} & \textbf{6.13} & \textcolor{lightgray}{-} \\\hline
    \end{tabular}
  \label{tab_exp_common_results_}
\end{table}

\subsection{Evaluating TCR on All Landmarks} \label{sec_exp_overall_}

Our evaluation in Section~\ref{sec_exp_common_} was confined to common landmarks between source and target.
Due to the unique capability of standardizing the annotation spaces between datasets, our model can readily exploits additional independent sources to enrich the annotations of a target dataset. 
To evaluate such capability, in this experiment we examined performance of our algorithm over full source-type annotations.
Similar to Section~\ref{sec_exp_common_}, we compared our model with the `closed-world' model (using SDM~\cite{Xiong_CVPR13}).

\begin{figure}
\centering
\includegraphics[width= \linewidth]{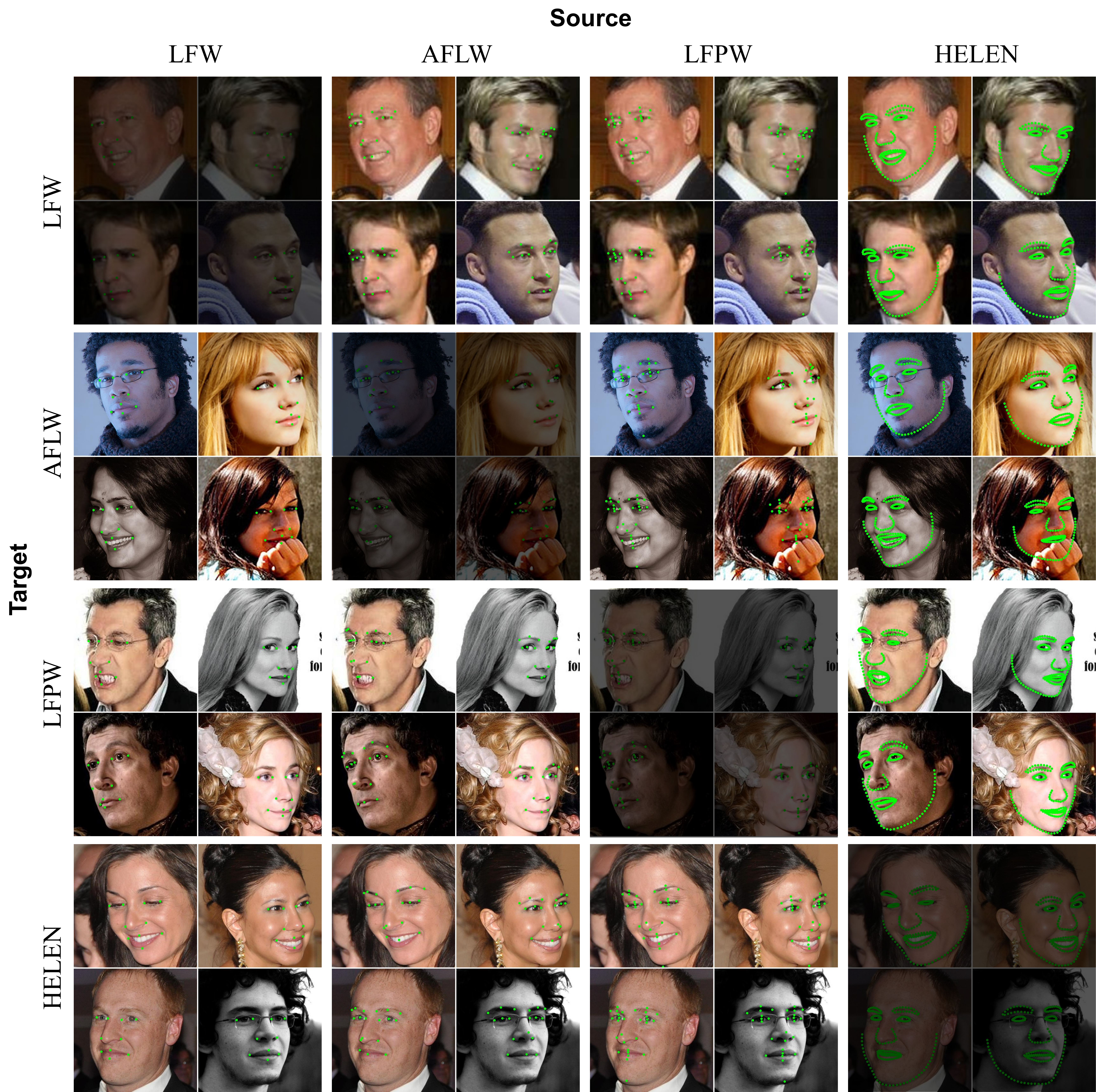}
\caption{Examples of face alignment results by the proposed Transductive Cascaded Regression (TCR) method. Each row shares the same testing target domain and each column shares the same source domain and same type of annotation.}
\label{fig_exp_overall_results_}
\end{figure}

Table \ref{tab_exp_overall_results_} summarizes the results\footnote{No results were reported on (HELEN / LFW-C) and (HELEN / AFLW-C) since we do not have the special tool for annotating 194 landmarks as in HELEN.}. It is observed that the TCR method outperforms the `closed-world' method in all cases.
Note that the proposed method employs target+source training data for learning a model, whilst the `closed-world' method only learns from the source training set. This is arguably an `unfair' comparison, but it highlights the importance of fusing different datasets for better generalisation. The fusion is not possible without the proposed transductive alignment approach.
Figure~\ref{fig_exp_overall_results_} shows the results by our transductive algorithm.

\begin{table}
  \centering
  \scriptsize
  \caption{Average error for all transferred landmarks. The right most column shows the relative improvement of the TCR against the `closed-world' method. $S=$ source, $T=$ target.}
  \label{tab_exp_overall_results_}
    \begin{tabular}{ l |c c c c || c c c c || c c c c }
    \hline
    \multicolumn{1}{r|}{$S$} & \multicolumn{4}{c||}{Closed-World (SDM \cite{Xiong_CVPR13})} & \multicolumn{4}{c||}{The Proposed TCR} & \multicolumn{4}{c}{Relative Improvement}\\
     \multicolumn{1}{l|}{$~T$} & LFW  & AFLW & LFPW & HELEN & LFW  & AFLW & LFPW & HELEN  & LFW  & AFLW & LFPW & HELEN \\\hline
    LFW-C & \textcolor{lightgray}{-} & 7.75 & 7.46 & - & \textcolor{lightgray}{-} & 7.11 & 7.36 & - & \textcolor{lightgray}{-} & 8\% & 1\% & - \\\hline
    AFLW-C & 6.62 & \textcolor{lightgray}{-} & 9.33 & - & 5.76 & \textcolor{lightgray}{-} & 8.81 & -  & 13\% & \textcolor{lightgray}{-} & 6\% & - \\\hline
    LFPW & 6.12 & 6.82 & \textcolor{lightgray}{-} & 5.97 & 3.97 & 5.82 & \textcolor{lightgray}{-} & 5.47 & 35\% & 15\% & \textcolor{lightgray}{-} & 8\% \\\hline
    HELEN & 8.02 & 8.38 & 6.65 & \textcolor{lightgray}{-} & 4.90 & 6.47 & 5.46 & \textcolor{lightgray}{-} &39\% &23\% & 18\% & \textcolor{lightgray}{-} \\\hline
    \end{tabular}
\end{table}
\normalsize

%

\subsection{Evaluating TCR on Unseen Samples with Occlusions} \label{sec_exp_cofw_}


In previous experiments we focused on training a model using target training set and have it tested on target testing set. In this experiment, we evaluated our model in an unseen domain with many samples have faces being partially occluded.
We selected the challenging Caltech Occluded Faces in the Wild (\textbf{COFW})~\cite{Burgos-Artizzu_ICCV13} for our evaluation.
The training set of COFW consists of nearly all training samples in LFPW and 507 extra faces with heavy occlusions. Its testing set contains 500 challenging samples with occlusion. The annotation type is identical to LFPW, with 29 landmarks annotated on each face.
To increase the challenge, we did not train our model on COFW, but the combination of LFPW and AFLW after applying the transductive alignment. Here, we used LFPW as source training data and AFLW as target training data.
We compared our method against RCPR~\cite{Burgos-Artizzu_ICCV13}, which is trained on COFW and specifically designed for handling faces with heavy occlusion. We used the publicly available implementation with the same parameter settings.

\begin{figure}
\centering
\includegraphics[width=\linewidth]{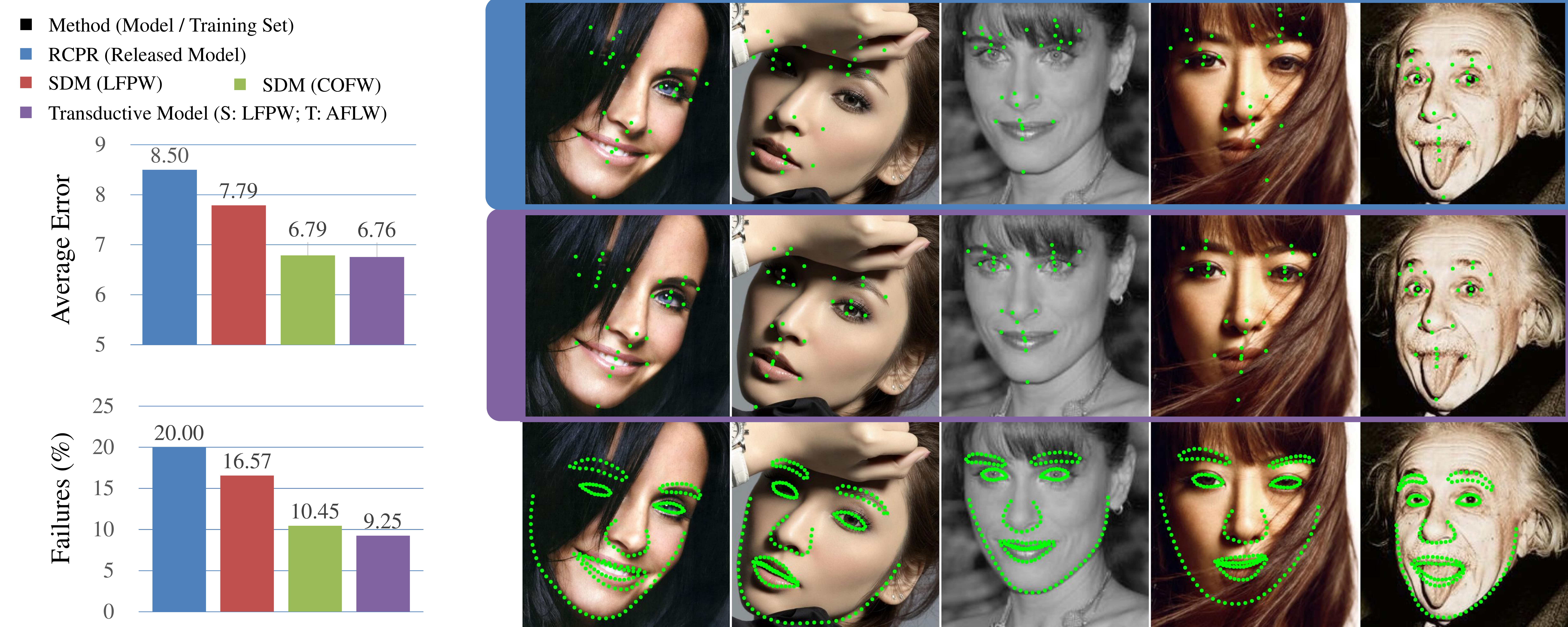}
\caption{Average error (\% interocular distance) and failure rate on COFW. The histogram on the left shows testing result on COFW when training on various training datasets with different methods. Examples on the right show the performance of our method compared with~\cite{Burgos-Artizzu_ICCV13}. The last row shows our results with HELEN-style annotations with AFLW as target domain. We did not evaluate the performance due to the lack of HELEN-style ground truth on COFW.}
\label{fig_exp_cofw_results_}
\end{figure}

The quantitative results and qualitative examples were summarized in Fig.~\ref{fig_exp_cofw_results_}. As expected, our model performs much better than the `closed-world' SDM method, which was trained only on source data, \ie~LFPW.
Interestingly, the proposed model, which are trained on LFPW and AFLW but without COFW, achieves competitive even better result than SDM with COFW as training set.
The results suggest the effectiveness of our model in combining the source LFPW with the target AFLW, leading to superior generalization even without the dedicated COFW as training set.

\section{Conclusion} \label{sec_conclusion}

%
We have formulated a novel Transductive Cascaded Regression (TCR) method, of which the core a transductive alignment approach, which is capable of transferring annotation style from one dataset to another seamlessly. Effectively bridging the annotation space allows one to combine two different datasets with diverse characteristics.
We have shown that a model trained on combined datasets performed extremely well in cross-dataset evaluation and even unseen domain with severe occlusion. In particular, our method has achieved \textbf{16.6}\% improvement on average against the `closed-world' method when performing cross-datasets evaluations, and \textbf{11.4}\% improvement on average compared to na\"{i}ve training sets fusion.
%

\bibliographystyle{splncs03}
\bibliography{ref_alignment_new}

\end{document}